\documentclass[letterpaper, 11pt]{article}
\usepackage[utf8]{inputenc}
\usepackage[margin = 1in]{geometry}
\usepackage[colorlinks, linkcolor=orange, anchorcolor=blue, citecolor=blue]{hyperref}
\usepackage{graphicx} 
\usepackage{xcolor}
\usepackage{siunitx}
\usepackage{indentfirst}
\usepackage[numbers]{natbib}
\usepackage{enumitem}
\usepackage{multirow}

\usepackage{listings}
\usepackage[noframe]{showframe}
\usepackage{framed}
\usepackage{lipsum}
\renewenvironment{shaded}{%
  \MakeFramed{\advance\hsize-\width \FrameRestore\FrameRestore}}%
 {\endMakeFramed}
\definecolor{shadecolor}{gray}{0.75}

\title{CRISPR-GPT for Agentic Automation of Gene-editing Experiments}

\date{}

\author{
  Yuanhao Qu$^{1,*}$, Kaixuan Huang$^{2,*}$, Ming Yin$^{2}$, Kanghong Zhan$^{3}$, \\ Dyllan Liu$^{4}$, Di Yin$^{1}$, Henry C. Cousins$^{5,6}$, William A. Johnson$^{1}$, \\ Xiaotong Wang$^{1}$, Mihir Shah$^{5}$, Russ B. Altman$^{4,7}$, Denny Zhou$^{8}$, \\ Mengdi Wang$^{2}$, Le Cong$^{1}$ \\ \\ 
}

\begin{document}

\maketitle
\def\thefootnote{$*$}\footnotetext{These authors contributed equally to this work. Correspondence to: \texttt{mengdiw@princeton.edu} (M.W.), \texttt{congle@stanford.edu} (L.C.). $^1$Department of Pathology, Department of Genetics, Cancer Biology Program, Stanford University School of Medicine, Stanford, CA, USA. $^2$Center for Statistics and Machine Learning, Department of Electrical and Computer Engineering, Princeton University, Princeton, NJ, USA. $^3$Department of Computing, Data Science, and Society, University of California, Berkeley, Berkeley, CA, USA. $^4$Department of Computer Science, University of California, Berkeley, Berkeley, CA, USA. $^5$Department of Medicine, Stanford University School of Medicine, Stanford, CA, USA. $^6$Medical Scientist Training Program, Stanford University School of Medicine, Stanford, CA, USA. $^7$Department of Bioengineering, Department of Genetics, Stanford University, Stanford, CA, USA. $^8$Google DeepMind, Mountain View, CA, USA. }

\begin{abstract}
    The introduction of genome engineering technology has transformed biomedical research, making it possible to make precise changes to genetic information. However, creating an efficient gene-editing system requires a deep understanding of CRISPR technology, and the complex experimental systems under investigation. While Large Language Models (LLMs) have shown promise in various tasks, they often lack specific knowledge and struggle to accurately solve biological design problems. In this work, we introduce CRISPR-GPT, an LLM agent augmented with domain knowledge and external tools to automate and enhance the design process of CRISPR-based gene-editing experiments. CRISPR-GPT leverages the reasoning ability of LLMs to facilitate the process of selecting CRISPR systems, designing guide RNAs, recommending cellular delivery methods, drafting protocols, and designing validation experiments to confirm editing outcomes. We showcase the potential of CRISPR-GPT for assisting non-expert researchers with gene-editing experiments from scratch and validate the agent’s effectiveness in a real-world use case. Furthermore, we explore the ethical and regulatory considerations associated with automated gene-editing design, highlighting the need for responsible and transparent use of these tools. Our work aims to bridge the gap between beginner biological researchers and CRISPR genome engineering techniques, and demonstrate the potential of LLM agents in facilitating complex biological discovery tasks. \textbf{The published version of this draft is available at \url{https://www.nature.com/articles/s41551-025-01463-z}.}
\end{abstract}

\section{Introduction}
Gene editing technology represents a groundbreaking scientific advancement that enables precise alterations to the genetic material of living organisms. This innovative technique has found extensive applications across various fields of biology and medicine, from correcting genetic defects responsible for disorders like cystic fibrosis, hemophilia, and sickle cell anemia, to offering new strategies in the battle against complex conditions such as cancer, cardiovascular diseases, neurodegenerative disorders, and infections. A most well-known gene editing system is called CRISPR-Cas9~\citep{n1_doudna2014new, n2_mali2013cas9, n3_hsu2014development, n4_sander2014crispr, n5_anzalone2020genome, n6_pickar2019next, n7_shalem2015high}. It was adapted from a naturally occurring genome editing system that bacteria use as an immune defense. Beyond CRISPR-Cas9, recent advancements have led to the development of CRISPR activation/interference, CRISPR-based prime editing and base editing technologies. CRISPR activation/interference, also called CRISPRa/CRISPRi, are able to either enhance gene expression or silence the activities of specific genes via epigenetic regulation~\citep{5_qi2013repurposing, 6_gilbert2013crispr, 7_konermann2013optical, 8_maeder2013crispr, 9_perez2013rna}. Prime editing, considered a "search and replace" method for DNA, allows for precise editing capabilities without introducing double-stranded breaks~\citep{10_anzalone2019search}. Base editing, on the other hand, enables the direct, irreversible conversion of one DNA base into another at targeted locations, further expanding the toolkit for precise genome modification~\citep{11_gaudelli2017programmable}. All these technologies hold potential for broad applications in medicine, agriculture, and beyond, enhancing the scope of genome editing in pursuing treatments for genetic diseases and other applications.

\begin{figure}[t]
    \centering
    \includegraphics[width=0.9\textwidth, bb=0 0 1002 543]{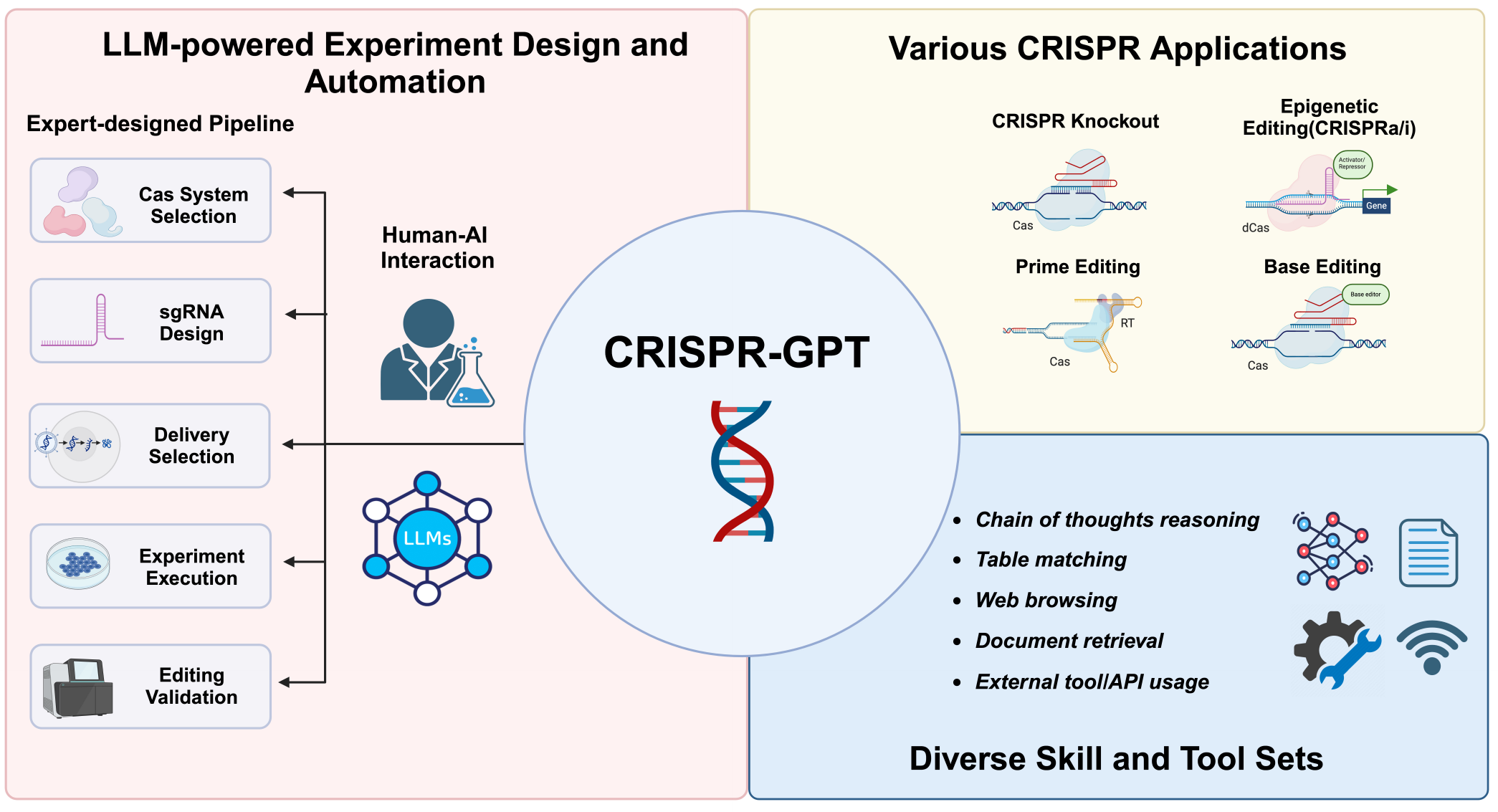}
    \caption{\textbf{Overview of CRISPR-GPT Agent}. The CRISPR-GPT is built upon an LLM-powered design and planning engine (left), which helps to complete 4 core meta-tasks (top right), as well as other auxiliary functions (freestyle Q\&A, off-target prediction). CRISPR-GPT has integrated a set of useful skills and toolkits (bottom right) that the LLM agent would call when needed to facilitate human users across the different tasks and subtasks. Figure created with BioRender.com.}
    \label{fig:1}
\end{figure}

Designing gene-editing experiments requires deep understanding of the suite of technologies and the associated biology of target organs.  CRISPR Cas-based editing works by interacting with the RNA of a short "guide" sequence (guide RNA) that binds to a specific target sequence in a cell's DNA, much like the RNA segments bacteria produce from the CRISPR array. When introduced into cells, the guide RNA recognizes the intended DNA sequence, and the Cas enzyme (often Cas9 or others) cuts the DNA at the targeted location, mirroring the process in bacteria. There are numerous considerations when designing such experiments, including the selection of a well-suited gene-editing system, the development of optimal guide sequences, and validation approach. This typically requires significant domain expertise, understanding of biology of the target organ, as well as trial-and-error efforts. Developing AI-assisted computation tools to aid gene-editing holds great promise to make the technology more accessible and accelerate both scientific and therapeutic developments.

Large language models (LLMs) have demonstrated exceptional capabilities in language skills and encapsulate a tremendous amount of world knowledge, approximating aspects of artificial general intelligence~\citep{12_chowdhery2023palm, 13_hoffmann2022training, 14_achiam2023gpt, 15_claude3, 16_anil2023palm}. Recent research has also explored enhancing LLMs with external tools, improving their problem-solving abilities and efficiency~\citep{17_yao2022react, 18_liu2022mind, 19_schick2024toolformer}. LLMs have also demonstrated potential as tool makers~\citep{20_cai2023large} and black-box optimizers~\citep{21_yang2023large}. Researchers have explored LLM-based specialized models for various application domains~\citep{22_li2024cancergpt, 23_wu2023bloomberggpt}, as well as for solving scientific and mathematical tasks. For example, ChemCrow~\citep{24_bran2023chemcrow} uses tool-augmented LM for solving a range of chemistry-related tasks such as paracetamol synthesis, whereas Coscientist~\citep{25_boiko2023autonomous}, also driven by GPT-4 and integrated automated experimentation, achieved successful optimization of palladium-catalyzed cross-coupling reaction.

\subsection{General-purpose LLMs do not know how to design biological experiments.}

While leveraging large language models (LLMs) for aiding in the design of gene-editing experiments presents an enticing prospect, the current state-of-the-art general-purpose models exhibit significant shortcomings in this specialized domain. These models, despite their vast knowledge base, lack the precise, up-to-date domain-specific knowledge essential for the accurate design of biological experiments. 

A critical limitation of general-purpose LLMs is their propensity for "hallucinations" or generating confident yet inaccurate responses when tasked with specialized biological queries. For instance, when tasked with designing a guide RNA (gRNA) sequence for targeting specific human genes, such as EMX1 or EGFR, general-purpose LLMs like ChatGPT-3/ChatGPT-4 can respond incorrect sequence with high confidence. However, the gRNA sequences they provide often do not correspond to any known genomic region. This discrepancy can be readily identified by comparing the LLM-generated sequences with reference sequences in databases such as the NCBI's BLAST tool, which aligns sequences to the human genome and transcriptome. Such hallucinated design sequences not only lack utility but can mislead researchers, potentially leading to wasted resources and time if not properly vetted.

In addition, responses generated from general-purpose LLMs usually lack essential details necessary for the experimental design, such as specific materials, protocols, considerations for off-target effects, gRNA efficiency, and specificity. Such gaps in information can leave researchers, especially those new to the field of gene editing, unprepared for the practical execution of experiments.

Furthermore, it is crucial to note that responses generated may contain overwhelming information that does not directly contribute to the gene-editing experimental designs. Such irrelevant text can lead to confusion and misdirection, complicating the researcher's task of identifying the most pertinent and practical information for their gene-editing objectives. 

All of these limitations underscore the necessity for a new class of LLMs tailored specifically for the gene-editing experimental designs (We refer readers to Appendix~\ref{sec:appendix:A} for more examples of failures). Such models would need to integrate deep, accurate domain knowledge with the ability to critically evaluate and generate experimentally viable solutions, thereby overcoming the current barriers faced by general-purpose LLMs in the design of CRISPR gene-editing experiments.

\subsection{Overview of CRISPR-GPT}

In the rapidly evolving field of genetic engineering, CRISPR technology has become a pivotal tool for precise gene editing. Despite its promise, the intricacy of designing CRISPR experiments—from guide RNA (gRNA) selection to predicting off-target effects—presents significant challenges, especially to those new to the field. To bridge this gap, we introduce CRISPR-GPT, a novel solution that combines the strengths of Large Language Models (LLMs) with domain-specific knowledge and computational tools, specifically tailored for CRISPR gene editing tasks.

CRISPR-GPT is centered around a tailor-made LLM-powered design and planning agent (Figure~\ref{fig:1}). This engine of the agent not only draws upon expert knowledge from leading practitioners in gene editing but also integrates a broad review of recent literature and a suite of computational toolkits including guideRNA design tool.

The innovation of CRISPR-GPT Agent allows automated designs of gene-editing experiments by simplifies the complex process into a series of manageable steps:
\begin{itemize}[itemsep=1pt, parsep=1pt, topsep=2pt]
    \item Selection of CRISPR System: Tailoring the choice of CRISPR system to the experiment's needs.
    \item gRNA Design: Optimizing guide RNA sequences for efficiency and specificity based on the Broad Institute’s gold-standard guideRNA library and CRISPRPick toolkit, including pre-designed gRNA libraries~\citep{26_kim2018deep, 27_deweirdt2021optimization,28_doench2016optimized, 29_sanson2018optimized}.
    \item Delivery Approach Selection: Advising on the most effective methods to introduce the CRISPR components into target cells.
    \item Prediction of Off-target Effects: Assessing potential unintended alterations alongside desired edits.
    \item Recommendation of Experimental Protocols: Outlining step-by-step procedures tailored to the experiment's objectives.
    \item Validation Approach Recommendation and Primer Design: Recommending gest ways to validate the edits and help design the associated primers. 
\end{itemize}
This approach, leveraging a chain-of-thought reasoning model and state machines, ensures that even individuals new to gene editing can iteratively refine their experimental designs to achieve protocols that meet their specific research needs. In addition, CRISPR-GPT offers:
\begin{itemize}[itemsep=1pt, parsep=1pt, topsep=2pt]
    \item A Freestyle Q\&A Mode for addressing ad hoc queries with precision,
    \item An Off-target Prediction Mode for in-depth analysis of pre-designed gRNAs.
\end{itemize}
These functions help the users when they meet additional issues during the experimental design process.

Mindful of the ethical and safety considerations surrounding gene editing, especially in human applications, we have integrated safeguards into CRISPR-GPT. These include restrictions on its use in human subjects, measures to ensure the privacy of genetic information, and alerts for potential unintended consequences, reflecting our commitment to responsible use in alignment with the broader scientific and ethical discourse on gene editing technologies.

\section{Methods and Algorithms}

\subsection{Large Language Model}

\begin{figure}[t]
    \centering
    \includegraphics[width=0.9\textwidth, bb=0 0 905 517]{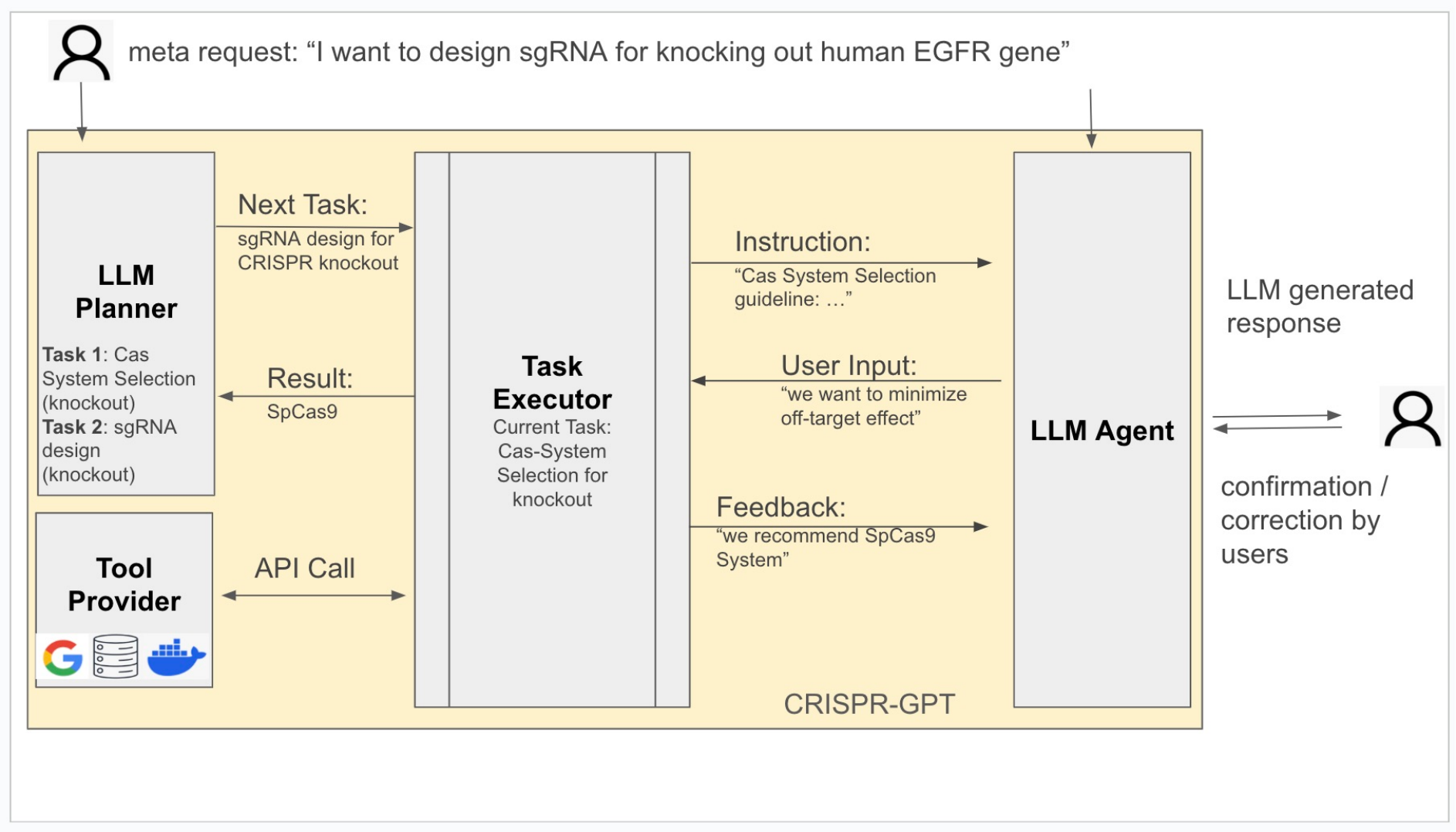}
    \caption{\textbf{Components of CRISPR-GPT enable human-AI collaboration to automate gene-editing experimental designs across complex tasks.} \textbf{LLM Planner} is responsible for configuring tasks based on the user’s needs (4 predefined meta-tasks or LLM planned chain of tasks). \textbf{Tool Provider} connects the system to external APIs, tools, libraries, and documents. \textbf{Task Executor} is implemented as a state machine, responsible for providing instructions and feedback, receiving input from LLM Agent, and calling APIs via Tool Provider. \textbf{LLM Agent} is responsible for interacting with the task executor on behalf of the user, where the user can monitor the process and provide correction to the LLM agent if the generated content.}
    \label{fig:2}
\end{figure}

The CRISPR-GPT agent consists of the following 4 core modules (Figure~\ref{fig:2}): LLM planner, Tool provider, Task executor, and the LLM Agent that serve as the interface with users for taking inputs and communicate outputs.

\subsubsection*{Task Executor operates as state machines, providing robust subgoal decomposition and progress control. } 

We implement 22 tasks, summarized in Table~\ref{tab:1}, in the form of state machines for CRISPR-GPT. The state machines are responsible for providing sufficient instruction for the current task and guiding the user to fulfill the decision-making through multiple rounds of textual interactions. Through these state machines, we manually decompose each task into sub-goals for the task executor. Specifically, each state is responsible for one particular sub-goal. The transition logic is well-defined so the task executor can properly transit to another sub-goal based on the current progress.

We have 4 predefined Meta-Tasks that support the full pipelines of 4 gene-editing-related experiments; see Table~\ref{tab:1}. Besides, the LLM planner can generate a customized list of tasks depending on the user’s meta-request. The state machines of the corresponding tasks are chained together as a bigger state machine to support the entire pipeline. 

\subsubsection*{Tool Provider connects Task Executor with external APIs}

To connect language models with external functionalities~\citep{30_chatgptplugin, 31_shen2024hugginggpt, 32_wang2023voyager, 33_hao2024toolkengpt, 34_patil2023gorilla}, the system needs to (1) analyze the current situation and judge whether it’s suitable to call an external tool; (2) know what kinds of tools are available and choose the best from them. Instead of directly exposing the interfaces of the APIs to LLMs, in CRISPR-GPT, we wrap the usage of APIs inside the states and expose more user-friendly and LLM-friendly textual interfaces through hand-written instructions and responses. In plain words, we are teaching users (human agents \& LLM agents) to use the tools. The tools include Google web search, running programs like Primer3~\citep{35_untergasser2012primer3}, as well as retrieval from external guide RNA libraries, research papers, and experiment protocols.   

\subsubsection*{LLM-planner automatically generates a list of tasks based on the user’s request}

\begin{figure}[t]
    \centering
    \includegraphics[width=0.9\textwidth, bb=0 0 928 526]{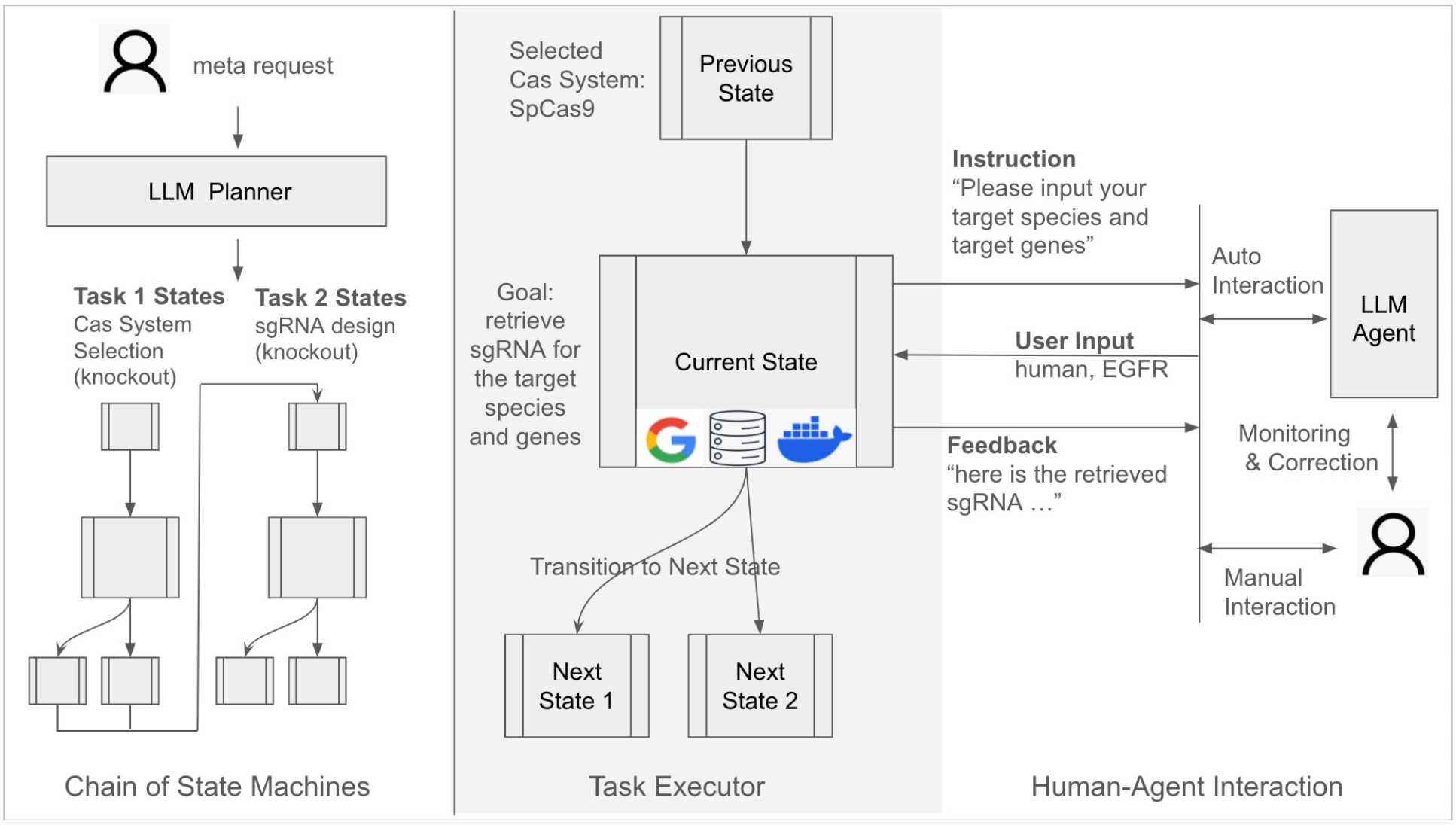}
    \caption{\textbf{Task decomposition process and state machine implementation algorithm.} \textbf{(Left) Task decomposition}; The LLMs can automatically perform task decomposition based on the user’s request, the descriptions of the current supported tasks and the dependencies, and the internal knowledge inside the LLMs. The state machines of the selected tasks are chained together to fulfill the user’s request.
 \textbf{(Right) State machines \& LLM Agent;} \textbf{ State machines} are the core of the Task Executor, where each state is responsible for one round of interaction with the user. The instruction is provided to the user first with sufficient information for the current decision-making step and the required inputs. After receiving the response from the user, it provides output and feedback, where APIs (e.g. program execution/web search/database retrieval) are potentially called during the execution of the state. Afterward, the state machine transits to the next state. \textbf{LLM Agent} generates responses to every step of the state machine on behalf of the user. The user monitors the whole process and provides corrections if the generated content is wrong or overrides the LLM Agent and manually interacts with the Task Executor.
}
    \label{fig:3}
\end{figure}

Large Language Models (LLMs) such as GPT-4~\citep{14_achiam2023gpt}, Gemini~\citep{team2023gemini}, and Claude~\citep{15_claude3} can serve as the reasoning core of the LLM-powered agent to solve real-world decision-making problems. We adopt the popular ReAct~\citep{17_yao2022react} prompting technique, where the LLM is prompted to output the chain-of-thought~\citep{37_wei2022chain} reasoning path and the final action from the plausible action set (Figure~\ref{fig:3}). To let LLMs perform task decomposition~\citep{38_zhou2022least}, we provide a table of the descriptions and the dependencies of all the tasks as a prompt to the LLM. Based on LLM’s internal knowledge as well as our manually written descriptions of tasks and instruction of task decomposition, LLM can intelligently analyze the user’s request and decompose the user’s request into a sequence of tasks, respecting the dependencies of the tasks. After the decomposition, the corresponding state machines are chained together to complete all the tasks. The prompt format of the task decomposition can be found in Appendix~\ref{sec:appendix:B}.

For robustness, we do not allow LLMs to dynamically add/delete new tasks (new state machines) during the automatic execution. However, we believe this is an important step toward a more intelligent CRISPR-GPT version and leave this as future work.

\subsubsection*{LLM-Agent automatically interacts with the Task Executor based on the user’s meta request}

In addressing the complex challenge of automating CRISPR gene editing tasks, we conceptualize the problem through the lens of sequential decision-making. This perspective frames the interaction between the user and the automated system as a series of steps, each requiring precise decisions to progress towards the ultimate goal of experiment design and execution. Central to our system is the LLM-agent, which acts as an intermediary between the user and a state machine. This state machine is derived from an initial task decomposition step, effectively breaking down the gene editing process into a structured sequence of actions and decisions. At each step in this sequence, the state machine presents a current state to the LLM-agent. This state encapsulates a description of the task at hand and specifies any input required from the user to move forward.

The LLM-agent's role is to interpret the current state and make informed decisions on behalf of the user. To do this effectively, the agent may draw upon a diverse set of information, including:
\begin{itemize}[itemsep=1pt, parsep=1pt, topsep=2pt]
    \item The instruction inherent to the current state,
    \item The specific request made by the user,
    \item A history of past interactions within the current task session,
    \item Results from external computational tools that have been integrated into the system.
\end{itemize}
This information is synthesized into a prompt for the LLM-agent, which then uses its capabilities to determine the most appropriate next action. The format and structure of these prompts, designed to optimize the decision-making process, are detailed in Appendix~\ref{sec:appendix:B}.

User oversight is a critical component of this system. While the LLM-agent operates autonomously, the user is not removed from the process. Instead, they are encouraged to monitor the progression of tasks and interact with the agent. This setup ensures that any errors or misinterpretations by the LLM-agent can be quickly identified and corrected by the user, maintaining the accuracy and integrity of the gene editing experiment design. This approach to automation emphasizes a collaborative synergy between human expertise and artificial intelligence. By leveraging the LLM-agent's ability to process and act on complex information, we facilitate a more efficient and user-friendly experience in designing CRISPR gene editing experiments. The sequential decision-making framework not only streamlines the task execution process but also ensures that user input remains a cornerstone of experiment planning and design.

\subsection{Human evaluation}

To evaluate the CRISPR-GPT agent's effectiveness in aiding gene editing and experimental design, we assembled a diverse group of 12 experts in the field of CRISPR and gene editing research. Each of the 12 experts rated the responses to the experimental design tasks from three modes on a scale from 1 (Poor) to 5 (Excellent) according to the established criteria (all rubrics for the human evaluation are detailed in Appendix~\ref{sec:appendix:C}). To offer a comparative perspective, outputs from both ChatGPT 3.5 and ChatGPT 4.0 (model version gpt-4-0613) were generated using similar prompts and evaluated using the same criteria.  

\subsection{Biological experiment and wet-lab validation}

We performed biological experiments through human-agent collaboration using CRISPR-GPT with ChatGPTv4 API, as wet-lab based real world validation of our approach. Specifically, we have independent scientist who are not familiar with gene-editing experiment to use CRISPR-GPT to facilitate their knock-out (KO) gene-editing experiments in a cancer research project. Detailed methods are provided below.

\textbf{Cell line and cell culture.}
A375 cell line was cultured in DMEM, high glucose, GlutaMAX (Gibco) supplemented with 10\% fetal bovine serum (FBS, Gemini Bio), 100 U/ml penicillin and 100ug/ml streptomycin (Gibco) at 37 $^\circ$C with 5\% CO2.

\textbf{crRNA cloning.}
Cloning of 4 crRNAs (TGFBR1/SNAI1/BAX/BCL2L1) was performed with BbsI or Esp3I (NEB) through a Golden Gate assembly approach into a Cas12a expressing backbone backbone. Constructs were sequence verified by Sanger sequencing using a U6 sequencing primer: 5’-GACTATCATATGCTTACCGT-3’.

\textbf{Lentivirus packaging and transduction.}
Lentivirus was produced by co-transfecting the assembled lentiviral vector with VSV-G envelope and Delta-Vpr packaging plasmids into HEK-293T cells using PEI transfection reagent (Sigma-Aldrich). Supernatant was harvested 48 hr after transfection. A375 cells were transduced at low MOI with 8\si{\micro\gram}/mL polybrene using a spin-infection at 1000*g for 45 minutes. After 24 hours, cells were selected with 1\si{\micro\gram}/mL puromycin to establish stably expressing cell lines.

\textbf{gDNA extraction, PCR and sequencing.}
Genomic DNA was extracted from selected cells 7 days later using QuickExtract (Lucigen). The targeted loci were then amplified using Phusion Flash High-Fidelity PCR Master Mix (ThermoFisher Scientific) according to the manufacturer’s instructions with primers containing Illumina sequencing adapters. Paired-end reads (150 bp) were generated on an Illumina MiSeq platform.

\section{Results}

CRISPR-GPT leverages the reasoning abilities of LLM, domain knowledge, retrieval techniques and external tools to provide a comprehensive solution to gene-editing experimental design tasks. It supports a wide arrange of gene editing scenarios, including single gene knockout, base editing without double strand breaks, insertions/deletions/replacement via prime editing, epigenetic editing for activation or repression of genes (CRISPRa and CRISPRi). 

\subsection{CRISPR-GPT assists researchers with gene-editing experimental design through three modules.}

\begin{figure}[h]
    \centering
    \includegraphics[width=0.6\textwidth, bb=0 0 1026 1189]{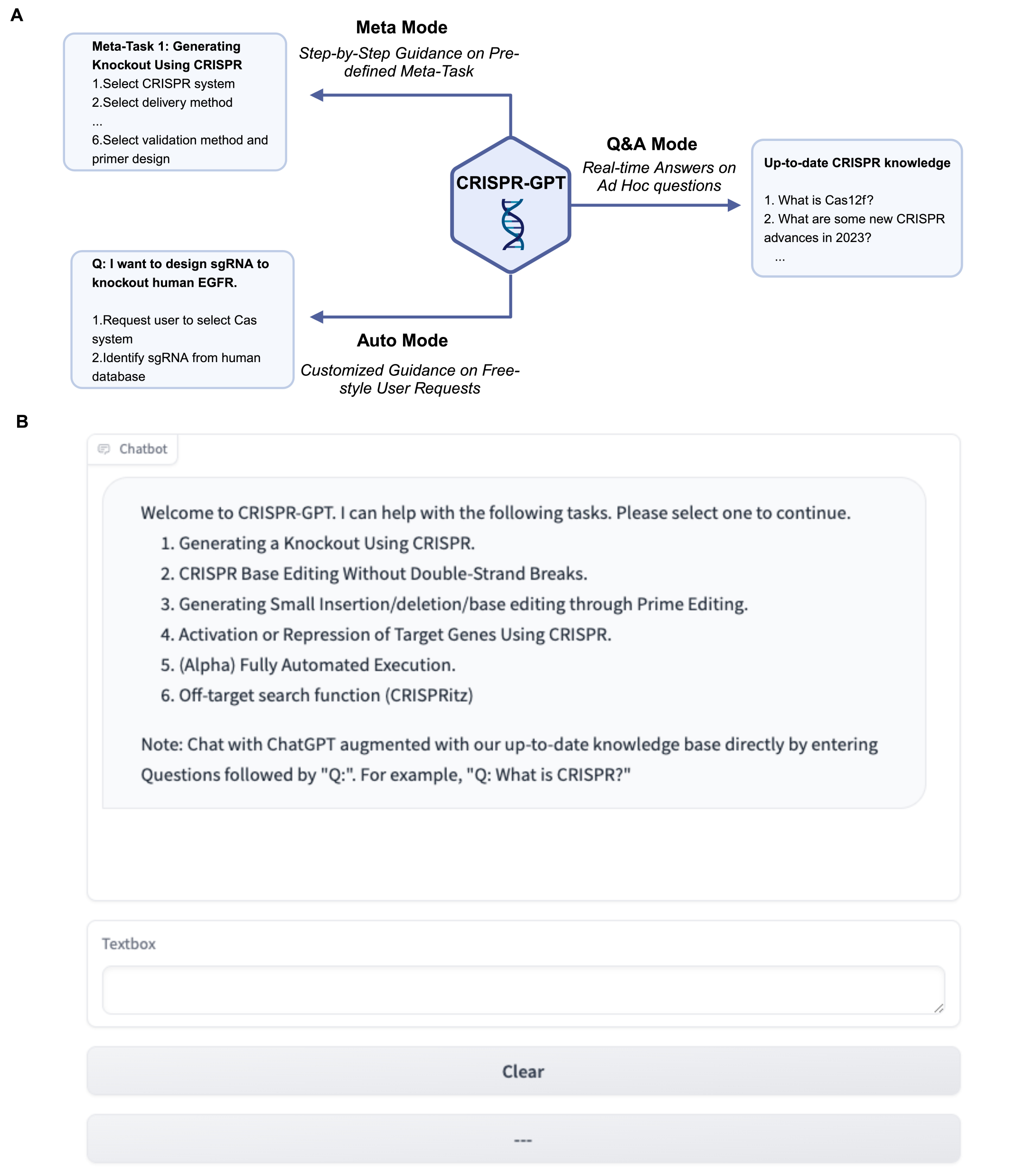}
    \caption{Overview of CRISPR-GPT’s interactive modules for gene-editing experimental design. \textbf{(A)} schematics illustrating the functionalities of the three modules within CRISPR-GPT, accompanied by examples of their applications. \textbf{(B)} Web-interface of CRISPR-GPT, note No.1-4 is the “Meta Mode”, No.5 is the “Auto Mode”, No.6 is Off-target-prediction function and “Q: prompt would trigger the Q\&A mode”.
}
    \label{fig:4}
\end{figure}

The CRISPR-GPT agent aids researchers in designing gene-editing experiments via three distinct modules. "Meta Mode" equips users, especially newcomers to the gene-editing field, with expertly defined pipelines for general gene editing scenarios (termed Meta-tasks). "Auto Mode" automatically generates a tailored list of necessary design tasks based on user input, facilitating goal achievement for users of all experience levels. "Q\&A Mode" functions as an advanced GPT-4 chatbot, addressing users’ CRISPR and gene-editing related queries throughout the design process (Figure~\ref{fig:4}).

\subsubsection{Meta Mode}

"Meta Mode" involves the planning and implementation of 22 unique gene-editing experimental design tasks utilizing four types of CRISPR-based gene editing systems (Meta-tasks) (Table~\ref{tab:1}). It leverages predefined pipelines to help users complete a Meta-task thoroughly. In this mode, the CRISPR-GPT agent guides users through each task necessary for designing gene editing experiments. This includes selecting the appropriate CRISPR system, recommending delivery methods, designing the sgRNA, predicting sgRNA off-target efficiency, selecting experimental protocols, and planning validation experiments.

\begin{figure}[h]
    \centering
    \includegraphics[width=0.9\textwidth, bb=0 0 1146 790]{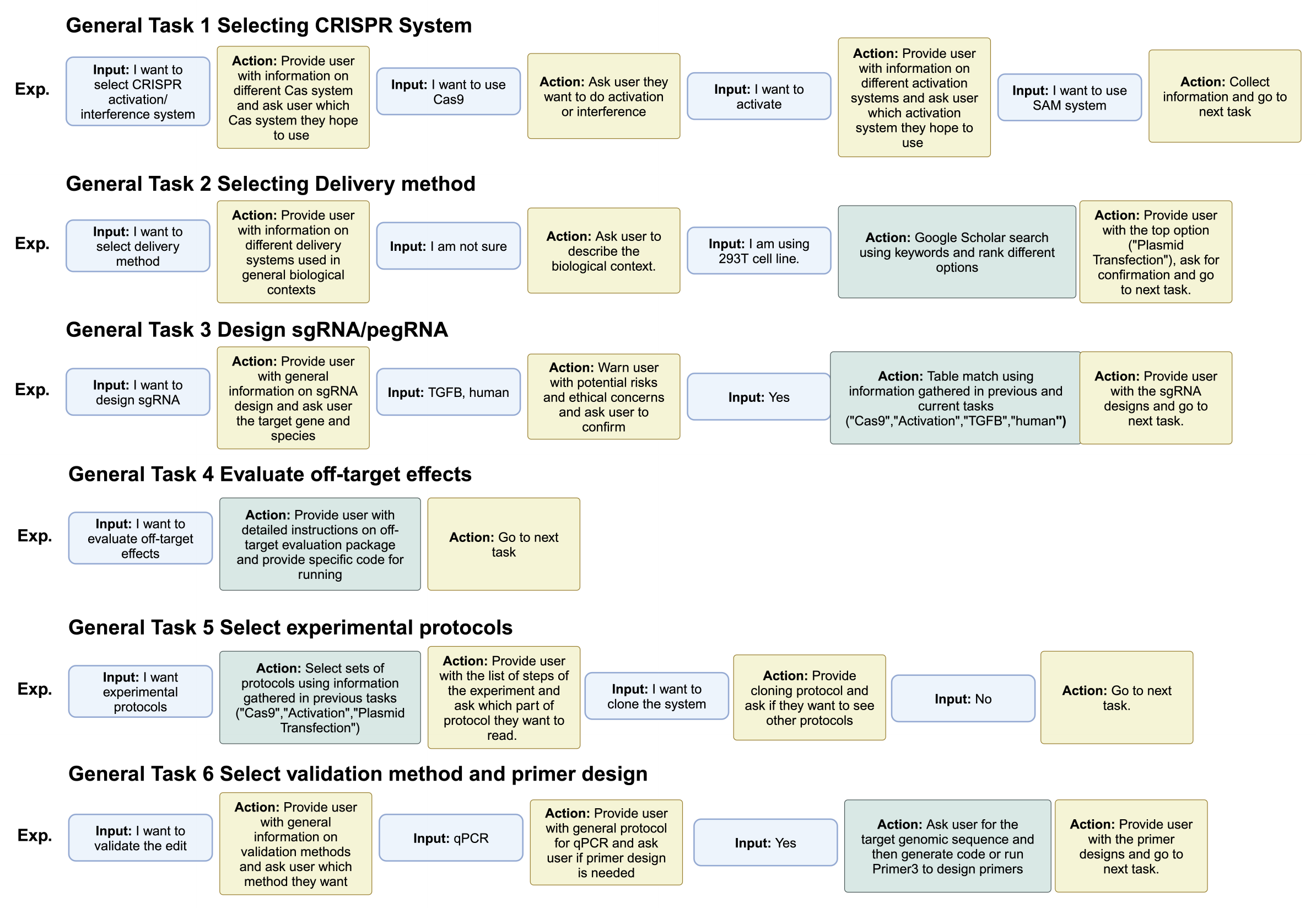}
    \caption{Example workflows outlining the general tasks involved in gene-editing experimental designs as facilitated by CRISPR-GPT.
}
    \label{fig:5}
\end{figure}

For every design task, the CRISPR-GPT agent interacts with users, applying various techniques and external tools to deliver the optimal solution. For example, in choosing CRISPR systems, CRISPR-GPT continuously interacts with users, providing instructions and collecting information to suggest options based on published protocols (See example in Figure~\ref{fig:5} General Task 1). For context-sensitive tasks like delivery method recommendations, CRISPR-GPT not only suggests common methods but also offers customized solutions based on the user’s requests through web search (See example in Figure~\ref{fig:5} General Task 2). For sgRNA/pegRNA designs, a multi-species database derived from existing designs and publications enables CRISPR-GPT to swiftly suggest pre-designed sgRNA based on user information (See example in Figure~\ref{fig:5} General Task 3). Following sgRNA/pegRNA design, users can evaluate designed guides' potential off-target effects with detailed instructions and code from CRISPR-GPT (See example in Figure~\ref{fig:5} General Task 4). After completing design tasks, CRISPR-GPT offers selected protocols based on the interaction history, including CRISPR system selection and delivery methods (See example in Figure~\ref{fig:5} General Task 5). Finally, for the validation task, CRISPR-GPT utilizes external APIs, like Primer3, to assist users in designing primers for validation experiments (See example in Figure~\ref{fig:5} General Task 6).

\begin{table}[]
\centering
\caption{List of meta-mode tasks (4 major meta-task and 22 specific tasks)}
\label{tab:1}
\begin{tabular}{|p{3cm}|p{5cm}|p{8cm}|}
\cline{1-3}
Meta-Tasks  & Gene editing scenarios     & Individual Design Tasks \\ \hline
\multirow{6}{3cm}{CRISPR Knockout} & \multirow{6}{5cm}{Single/multiple genes knockout, deletion of gene fragments} & CRISPR/Cas system selection~\citep{39_ran2013genome}  \\ \cline{3-3} 
                  &                   &  Delivery method selection \\ \cline{3-3} 
                  &                   & sgRNA design for knockout~\citep{26_kim2018deep, 27_deweirdt2021optimization, 28_doench2016optimized, 29_sanson2018optimized} \\ \cline{3-3}
                 &                    &   off-target evaluation~\citep{40_cancellieri2020crispritz} \\ \cline{3-3} 
                   &                    &  experimental protocol recommendation~\citep{41_giuliano2019generating}  \\ \cline{3-3}
                   &                    & validation protocol recommendation and primer design for sequencing~\citep{35_untergasser2012primer3,41_giuliano2019generating} \\ \hline
\multirow{6}{3cm}{CRISPR activation /interference} & \multirow{6}{5cm}{Gene activation and repression} & CRISPR/Cas Activation/Interference system selection~\citep{42_du2016crispr}  \\ \cline{3-3} 
                  &                   &  Delivery method selection \\ \cline{3-3} 
                  &                   & sgRNA design for activation/interference~\citep{26_kim2018deep, 27_deweirdt2021optimization, 28_doench2016optimized, 29_sanson2018optimized} \\ \cline{3-3}
                 &                    &   off-target evaluation~\citep{40_cancellieri2020crispritz} \\ \cline{3-3} 
                   &                    &  experimental protocol recommendation~\citep{41_giuliano2019generating}   \\ \cline{3-3}
                   &                    & validation protocol recommendation and primer design for qPCR~\citep{35_untergasser2012primer3,41_giuliano2019generating} \\ \hline
\multirow{6}{3cm}{CRISPR Base Editing} & \multirow{6}{5cm}{Single base replacement from CG to AT or AT to CG and broad mutagenesis} & Base editing system selection~\citep{43_huang2021precision}  \\ \cline{3-3} 
                  &                   & Delivery method selection \\ \cline{3-3} 
                  &                   & sgRNA design for base editing~\citep{44_hanna2021massively} \\ \cline{3-3}
                 &                    &   off-target evaluation~\citep{40_cancellieri2020crispritz} \\ \cline{3-3} 
                   &                    &  experimental protocol recommendation~\citep{41_giuliano2019generating} \\ \cline{3-3}
                   &                    & validation protocol recommendation and primer design for sequencing~\citep{35_untergasser2012primer3,41_giuliano2019generating}\\ \hline
\multirow{6}{3cm}{CRISPR Prime Editing} & \multirow{6}{5cm}{Small fragment insertion, replacement, and deletion} & Prime editing system selection~\citep{45_doman2022designing} \\ \cline{3-3} 
                  &                   & Delivery method selection \\ \cline{3-3} 
                  &                   & pegRNA design for prime editing\citep{25_boiko2023autonomous, 46_hwang2021pe, 47_chow2021web, 48_morris2021automated} \\ \cline{3-3}
                 &                    &   off-target evaluation~\citep{40_cancellieri2020crispritz}  \\ \cline{3-3} 
                   &                    &  experimental protocol recommendation~\citep{41_giuliano2019generating} \\ \cline{3-3}
                   &                    & validation protocol recommendation and primer design for sequencing~\citep{35_untergasser2012primer3,41_giuliano2019generating}\\ \hline

\end{tabular}
\end{table}

\subsubsection{Auto Mode}

"Auto Mode" also facilitates the planning and execution of 13 unique gene-editing experimental design tasks. Unlike "Meta Mode," it does not rely on predefined meta-tasks and pipelines; instead, it uses an LLM-planner to break down a user's request into a sequence of dependent tasks. For instance, if a user requests to “design sgRNA to knockout human EGFR,” the CRISPR-GPT agent identifies the keywords from the request and lists the necessary design tasks, like “CRISPR/Cas system selection” and “sgRNA design for knockout.” Additionally, it uses information from the initial request (e.g., target gene “EGFR” and species “human”) to autofill relevant fields and generate sgRNA designs without needing repeated inputs from the user. Simultaneously, CRISPR-GPT elucidates the rationale behind its choices, allowing users to track the process and make corrections if necessary.

\subsubsection{Q\&A Mode}

During the design tasks in "Meta Mode" and "Auto Mode," the CRISPR-GPT agent offers immediate responses or advice for CRISPR and gene editing related inquiries through "Q\&A Mode." For example, after selecting a CRISPR system, users seeking more information about the chosen system (e.g., Cas12a) can quickly obtain answers by asking, “Q: What is Cas12a?”. The CRISPR-GPT uses its knowledge base and document retrieval from expert-selected databases in the field to provide accurate and relevant information swiftly.

\subsection{CRISPR-GPT outperforms general LLMs in gene-editing design tasks through human expert evaluations.}

To evaluate the performance of CRISPR-GPT agent, we invited 12 researchers with expertise in CRISPR and gene editing to design sets of tasks to test the ability of CRISPR-GPT in assisting researchers with experimental design. The results are evaluated in four different aspects: Accuracy, Reasoning, Completeness and Conciseness (Appendix~\ref{sec:appendix:C}). Accuracy reflects whether CRISPR-GPT could provide accurate information of current state of CRISPR research and methodologies. Reasoning evaluates if CRISPR-GPT could provide insightful, well-supported explanation of the suggested designs. Completeness makes sure users receive all required information needed for CRISPR experimental design. Lastly, conciseness ensures that CRISPR-GPT provides users with directly relevant information for the design tasks with minimal unnecessary information. All evaluators were asked to score sets of tasks with these four aspects from 1 (Poor) to 5 (Excellent) for all three modes. Responses from ChatGPT 3.5 and ChatGPT 4.0 were generated and scored alongside those of CRISPR-GPT, using equivalent prompts in all cases.

\begin{figure}[h]
    \centering
    \includegraphics[width=0.9\textwidth, bb=0 0 734 288]{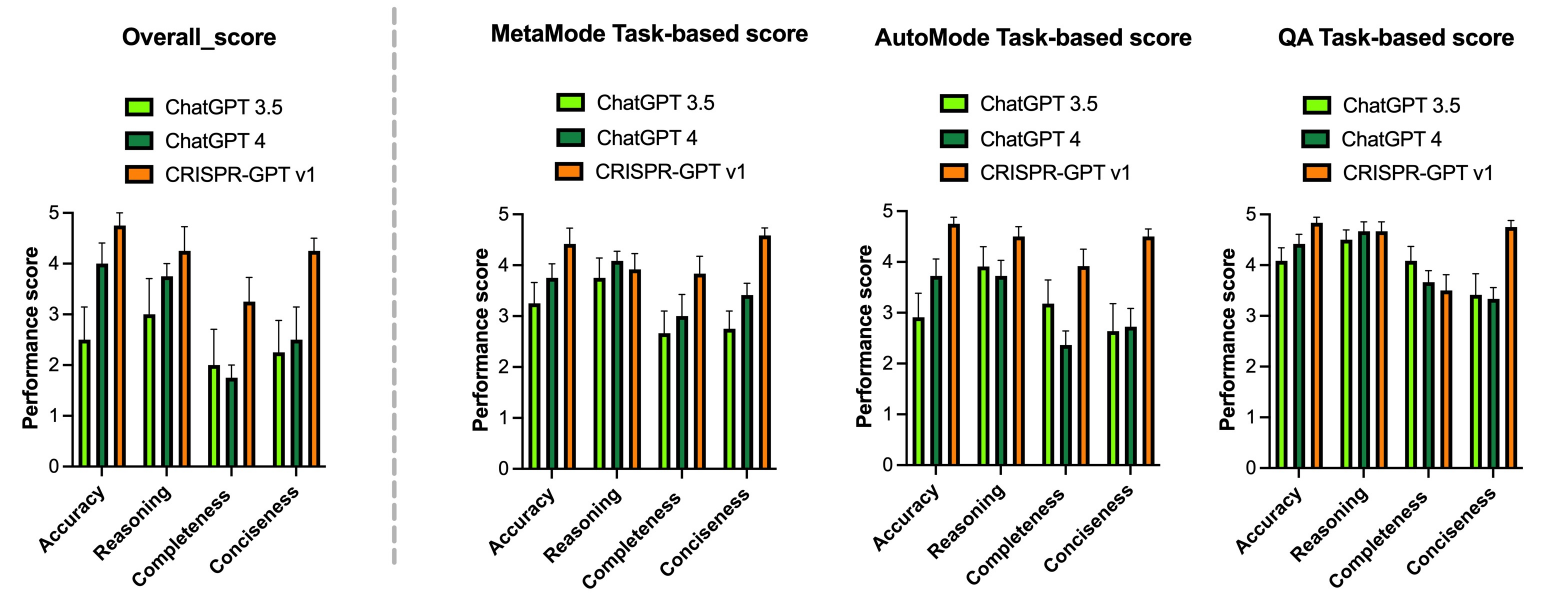}
    \caption{Evaluation results showing comparative performance of CRISPR-GPT and ChatGPT 3.5/4.0 in a range of gene-editing experiment design tasks across three different modes: MetaMode, AutoMode and QAMode. All tasks are scored in four different aspects: accuracy, reasoning, completeness, and conciseness. A score of 1 represents poor performance and 5 represents excellent performance. Detailed rubrics are listed in Appendix~\ref{sec:appendix:C}.
}
    \label{fig:6}
\end{figure}

We observed that CRISPR-GPT achieved significantly higher accuracy in our designed sets of tasks over general LLM-agents across all three modes, as we employed vast domain knowledge in the CRISPR and gene editing field to ensure the robustness of the CRISPR-GPT agent (Figure~\ref{fig:6}). While, responses generated by general LLM agents including ChatGPT 3.5 and ChatGPT 4.0 contain more minor factual errors due to known issues including inadequate domain knowledge and hallucination. At the same time, we found both CRISPR-GPT and general LLM agents show good reasoning ability over different sets of tasks. For “Auto Mode” related tasks, CRISPR-GPT showed even better reasoning, potentially due to better prompting techniques encoded in the agent. As we expected, “Completeness” is the major issue for general LLM-agents to perform gene-editing experimental design tasks. They can usually provide general guides for designs while could not provide design details due to the lack of domain knowledge and external tools. On the opposite side, CRISPR-GPT showed much better “Completeness” performance scores in the design tasks, allowing the users to perform the gene-editing experiment solely based on the information provided by CRISPR-GPT. It's worth noting that both ChatGPT 3.5 and 4.0 outperform CRISPR-GPT in the “Completeness” performance score in “Q\&A” mode. Such a result is due to the intentional tradeoff between “Completeness” and “Conciseness”. Answers directly generated by general LLM-agents usually include much irrelevant information in order to provide users with a more complete response. This usually confuses the users and makes it hard to catch the key information. In this case, we intentionally designed the CRISPR-GPT to provide concise accurate answers to the users across all different modes and accordingly CRISPR-GPT showed consistent better “Conciseness” performance scores.
 
Overall, through experts’ evaluation, we found CRISPR-GPT showed significantly improved performance over general LLM-agents for gene-editing experimental design tasks in all different aspects. Notwithstanding, CRISPR-GPT experienced difficulties in more complex gene editing scenarios and rare biological cases. It can be further extended and improved in the future with more up-to-date domain knowledge and better external tool sets.

\subsection{CRISPR-GPT demonstrates its efficacy through real-world application.}

\begin{figure}[h]
    \centering
    \includegraphics[width=0.9\textwidth, bb=0 0 1124 643]{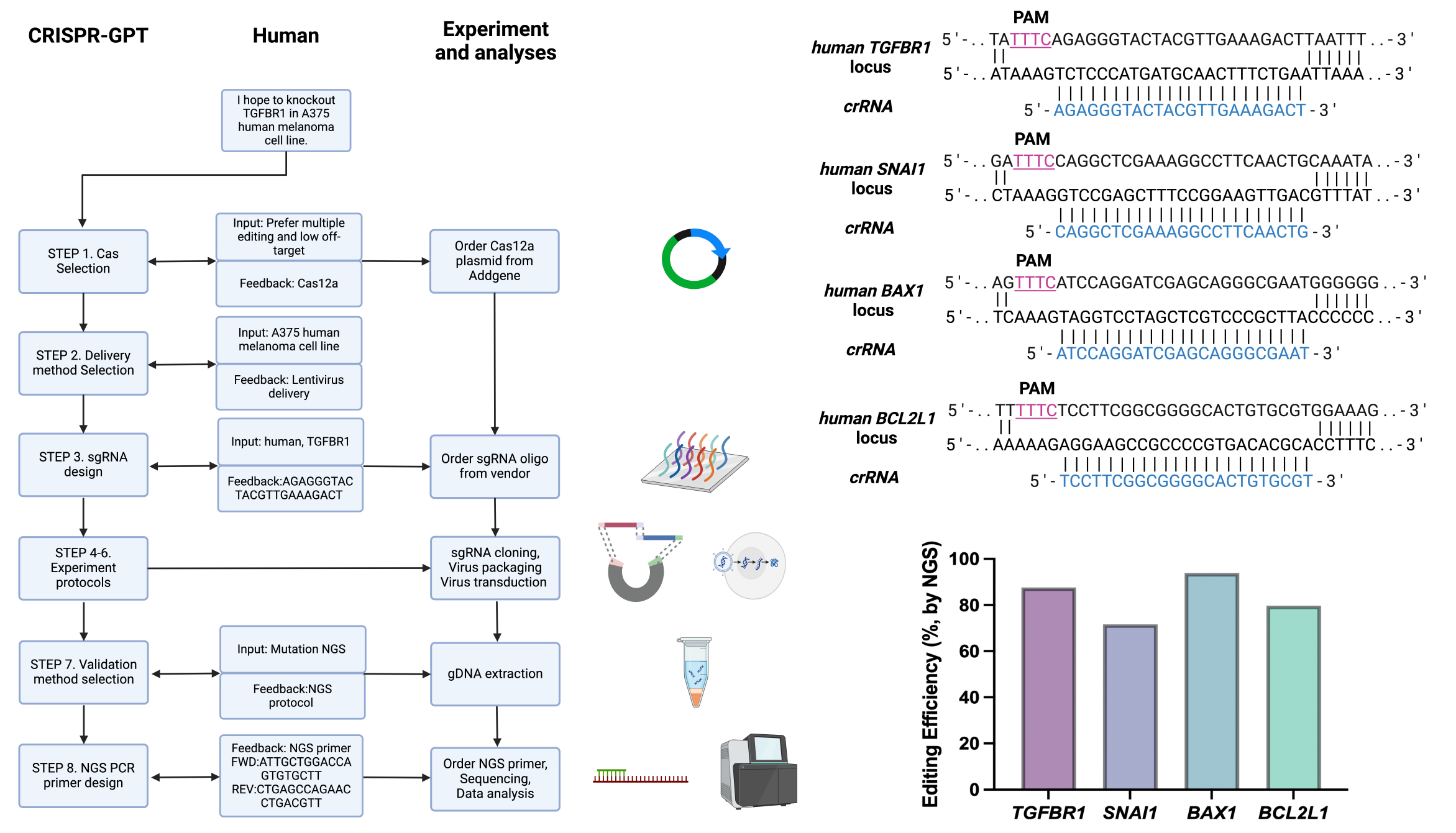}
    \caption{Wet-lab demonstration of human-AI collaboration in performing a gene-knockout experiment. \textbf{(Left)} Schematic showing the workflow of human-AI interaction. \textbf{(Top right)} Sample sgRNAs designed through CRISPR-GPT. \textbf{(Bottom right)} Editing outcome from next generation sequencing.
}
    \label{fig:7}
\end{figure}

To demonstrate CRISPR-GPT in assisting researchers with designing gene-editing experiments, we performed a gene knockout experiment in the human A375 cell line through continuous interaction with CRISPR-GPT (Figure~\ref{fig:7}).
 
In this experiment, we aimed to knock out a panel of 4 genes (TGFBR1, SNAI1, BAX, BCL2L1) individually in the human A375 cell line. To start, we selected the “Meta mode” to design the gene knockout experiment from scratch. Following the instruction of selecting the CRISPR system in the CRISPR-GPT, we selected AsCas12a as we hope to perform multiple-site edits and lower the potential off-target editing rate. For the delivery of the CRISPR system in A375 cells, we followed CRISPR-GPT's recommendation to use lentiviral transduction, ensuring stable expression of both the Cas enzyme and sgRNAs.
 
Then, based on this information, we were able to obtain the Cas12a plasmid (previously owned). When it came to designing the sgRNAs, we specifically targeted the human TGFBR1/SNAI1/BAX/BCL2L1 genes, fully aware of the ethical ramifications of human gene editing raised by CRISPR-GPT. CRISPR-GPT provided 4 sgRNA sequences for each gene (Exampled in Figure~\ref{fig:7}) from a published library so we were able to order the sequences for synthesis.

Later, CRISPR-GPT offered protocols for gRNA cloning. Detailed instructions were then provided for producing lentivirus through calcium phosphate transfection in HEK293T cells, using the necessary plasmids and viral packaging components. Following this, we exactly followed the protocol generated by CRISPR-GPT, through the transduction process, which involved cell culture procedures, the addition of lentivirus, and the use of polybrene to facilitate efficient transduction. To perform validation, we chose next-generation sequencing (NGS) for mutation detection and validation of the knockout in CRISPR-GPT, guided by the protocol provided by CRISPR-GPT agent. To prepare for NGS, we extracted genomic DNA from cells using the DNeasy Blood \& Tissue Kit based on the protocols. For the crucial step of PCR primer design, we provided detailed sequence information to CRISPR-GPT, which automatically returned a set of primers designed with Primer3 to specifically amplify the target site. In the concluding stages of our experiment, CRISPR-GPT advised us to attach Illumina adaptors to the PCR products for library construction and emphasized the necessity of checking primer specificity with NCBI BLAST. This final validation step was crucial to prevent mis-priming and to ensure that the sequencing results would accurately reflect the intended genomic edits.
 
Finally, we analyzed the data from the NGS, and observed a consistent high rate of expected editing outcomes across all 4 targeted genes. Through this process, where CRISPR-GPT provided: (1) CRISPR system selection (2) guideRNA design (3) delivery system recommendation (4) plasmid and viral vector selection with cloning protocol (5) tissue culture, cell transduction procedures (6) cell harvesting and gene-editing efficiency quantification methods (7) sequencing primer design and readout validation protocol. Thus, the dynamic interaction between our expertise and CRISPR-GPT's computational guidance was instrumental in executing a precise and ethically considerate gene-editing experiment.

\section{Safety and Ethical Concerns}

Safety and ethical concerns arise when using AI tools to guide genome editing, ranging from the risk of illegally altering human genomes to privacy issues when user genome information is involved. 

\subsection{Mitigation of the risk of human heritable editing}

Technologies such as CRISPR-Cas9, have made it possible to alter human genomes, which pose a number of ethical and safety risks. In particular, germline cell and embryo genome editing bring up a number of ethical challenges, including whether it would be permissible to use this technology to enhance normal human traits (such as height or intelligence). Based on concerns about ethics and safety, germline cell and embryo genome editing are currently illegal in the United States and many other countries. To ensure CRISPR-GPT follows the guidelines given in a moratorium~\citep{49_lander2019adopt} on heritable genome editing.

CRISPR-GPT employs a mechanism to make sure in all tasks users cannot bypass the existing step asking which organism they are editing. The agent would check if the editing target belonged to human tissues or organs. If it is found that the editing target is a human organ, it will trigger the following solution: Warning note when users proceed with designing human gene-editing exp. Link to this international moratorium with note. Ask users to confirm they understand the risk and have read this international guideline before proceeding.

\subsection{Protection of user genome data privacy}

Other concerns are related to user data privacy issues, especially when human genome sequence information might be exchanged by using AI tools. We follow the data privacy and HIPAA privacy rule in healthcare~\citep{50_hippa}. Although genome-scale sequences are fundamentally linked to identities, DNA segments of up to 20 bp length are considered safe and not able to identify human identity (REF). CPISPR-GPT is equipped with the following functionalities, to avoid supplying any identifiable private human/patient sequence to a public LLM model. Specifically, our solution is:
\begin{itemize}[itemsep=1pt, parsep=1pt, topsep=2pt]
    \item CRISPR-GPT would never store any identifiable long genome sequence in the server that would potentially reveal patient private information.
    \item CRISPR-GPT implements a filter to detect if there is any $\geq$ 20bp of A/T/G/C/U sequence contained in the prompts before sending them to external LLMs. After detecting the existence of such a sequence, the agent would raise an error with a warning note, asking the user to manually delete such sequences in the input. In this way, it avoids leaking such sensitive information to the public LLM model.
\end{itemize}

\section{Discussion}

The CRISPR-GPT agent showcases the remarkable potential of LLMs in automating and enhancing the design process of complex biological experiments. By seamlessly integrating LLMs with domain knowledge, external tools, and a modular task execution system, CRISPR-GPT empowers researchers to navigate the intricate landscape of CRISPR gene editing experiments with unprecedented ease and efficiency. The multi-modal capabilities of CRISPR-GPT encompass meta-task pipelines, interactive prompts, and on-demand Q\&A support. Researchers can leverage the agent's expertise to plan and execute gene editing experiments, from CRISPR system selection and guide RNA design to automated drafting of detailed protocol and validation strategies. This streamlined workflow not only accelerates the design process but also mitigates the risk of errors and oversights, thereby enhancing the quality and reproducibility of research outcomes.

While there exist LLM agent in other scientific domains such as chemistry, the complexities of biological experiments involving living materials demand a distinct set of considerations. Unlike chemical reactions, which often follow well-defined protocols, biological experiments require intricate procedures that account for the dynamic nature of living systems. CRISPR-GPT addresses this challenge by providing detailed, step-by-step guidance tailored to the specific experimental context, ensuring that researchers can navigate the nuances of working with living cells and organisms effectively.

Moreover, the free-style prompting and ad hoc Q\&A capabilities of CRISPR-GPT set it apart from many existing agents. Researchers can pose unstructured queries and receive contextualized responses, facilitating a more natural and intuitive interaction with the agent. This feature is valuable in the face of unexpected challenges or unforeseen circumstances that may arise during the course of an experiment, enabling researchers to seek timely guidance and adapt their approach as needed.

Despite its impressive capabilities, CRISPR-GPT is not without limitations. While the agent can design individual components, such as guide RNAs and primers, it currently lacks the ability to generate complete constructs or vectors from natural language input. This limitation highlights an area for future development. For example, recent advancements in the field of modular design of gene-editing, such as FragMID~\citep{51_mcgee2024modular} could be integrated with CRISPR-GPT to realize the potential for LLMs to empower researchers to explore and optimize CRISPR design and customized strategies, leading to more efficient gene-editing. 

Looking ahead, the integration of CRISPR-GPT with automated laboratory platforms and robotics holds immense promise. By bridging computational design and physical execution, researchers could leverage the agent's expertise to orchestrate end-to-end automated experiments, minimizing manual intervention and accelerating the pace of discovery.

\bibliography{ref.bib}
\bibliographystyle{plain}

\newpage
\appendix

\section{Failure Cases of GPT-4}
\label{sec:appendix:A}
The GPT version we use is gpt-4-0613.

\subsection{Examples of inaccurate information}

\subsubsection*{Example 1:}

\textbf{Prompt:} Which specific base editing system I should use to broadly introduce mutations to human EGFR gene?

\textbf{GPT4:}
If you are looking to induce a spectrum of mutations in the human EGFR gene to study the phenotypic implications of these mutations, utilizing multiple base editing systems would be beneficial.
1. **Adenine Base Editors (ABEs):** These can be used to induce A-G transitions in your target gene. ABEs are helpful as they induce minimal bystander mutations and indels compared to cytosine base editors (CBEs).
2. **Cytosine Base Editors (CBEs):** You can incorporate these when a C to T change (or G to A on the opposite strand) is desired in your target gene. Examples of CBEs include BE3, Target-AID, and BE4.
3. **Prime Editing:** If more diverse mutations beyond simple transitions are needed, or insertion/deletion mutations are needed, consider using prime editing. Although more complex, the versatility of prime editing allows it to create a wider range of mutations without inducing double-strand breaks or relying on homology-directed repair.
To ensure that your mutations are broad, careful design of guide RNAs for each specific location of your target gene where mutations are required is necessary. These gRNAs should be specific to the location in the target gene where A-G or C-T substitutions are desired.
However, take note that introducing broad mutations could also increase the possibility of off-target effects, hence it is crucial to have stringent control of your editing experiment, and thorough validation of the induced mutations by techniques such as Sanger sequencing or Next-Generation Sequencing (NGS).
As always, bear in mind the ethical considerations and regulations applicable to your region, as well as potential biosafety issues, especially when working with oncogenes such as EGFR.

\textbf{Note:} The answer fails to mention the specific expanded-window/non-specific base editors. It provides irrelevant information on ABE, CBE and prime editing.

\subsubsection*{Example 2:}

\textbf{Prompt:} Which one of the delivery method I should use to deliver CRISPR knockout systems in HepG2 cell line?

\textbf{GPT4:}
Lipofection is commonly used to deliver CRISPR knockout systems into HepG2 cells. Lipofectamine 3000 is a reagent often utilized for this purpose.
Remember to take into account that the transfection conditions might need to be optimized based on the specific characteristics of your experimental design. Always follow the manufacturer's guidelines related to the specific CRISPR system being used in order to ensure the highest level of efficiency.
If for any reason lipofection is not the preferable choice, electroporation or lentiviral transduction could be other options depending on the specifics of the experiment and resources available.

\textbf{Note:}
As HepG2 is a well-known hard-to-transfect cell line, GPT4 provides inaccurate information of using lipofection to deliver the CRISPR knockout system as the top choice.

\subsubsection*{Example 3:}

\textbf{Prompt:} How should I verify the CRISPR knockout of EGFR in the A549 cell line?

\textbf{GPT4:}
After performing a CRISPR-Cas9 mediated knockout of EGFR in A549 cells, you should confirm the knockout on a genotypic and phenotypic level. Below are some methods for assessing this:
1. PCR and Sequencing: Extract genomic DNA from the transfected cells and conduct PCR to amplify the region surrounding the CRISPR target site. Subsequent Sanger sequencing can identify any insertions or deletions (indels) at the target site.
2. T7 Endonuclease I Assay or Surveyor Nuclease Assay: These mismatch cleavage assays can also be used to detect the indels introduced by CRISPR-Cas9 activity.
3. Western Blot: To confirm loss of EGFR expression at the protein level, perform a Western blot using an anti-EGFR antibody. You should observe a lack of or significant reduction in the EGFR band in the CRISPR-treated cells compared to controls.
4. Immunofluorescence: Perform immunofluorescence staining for EGFR. The CRISPR-treated A549 cells should show lower fluorescence signal compared to control cells.
5. Cell Function Analysis: As the knockout of EGFR can impact EGFR downstream signaling and cell proliferation, the analysis of cell growth, migration, or signaling changes can provide further evidence of functional knockout. 
6. qRT-PCR: To assess EGFR mRNA expression levels, quantitative real-time PCR can be performed.
7. Next-Generation Sequencing (NGS): For a more detailed understanding of the specific mutations and possible off-target effects introduced by CRISPR, NGS can be performed.
Always include appropriate controls in your experiment and conduct the experiments in triplicates for statistical significance. It is generally recommended to use multiple verification methods to validate the results.

\textbf{Note:}
The answer provides wrong information of using qRT-PCR to validate the knockout status. It is known that qRT-PCR is not a reliable technique to assess gene knockout edits generated with a single guide RNA.

\subsubsection*{Example 4:}
\textbf{Prompt:} Could you give me a set of sgRNA sequences of Cas9 targeting human EGFR?

\textbf{GPT4:}
It's important to note that the efficiency and specificity of sgRNA sequences can depend on several factors, including the particular experimental conditions and target population of cells. Here are a few sgRNA sequences targeting human EGFR that have been used in published studies:
1. 5'- GTGGAGTGTGTTGAGCTTGG -3'
2. 5'- GCGTGCAGATGAACTTCTCAAGG -3'
3. 5'- GCTGGTCCTGGAGGAGGTCAAGG -3'
4. 5'- CCTGGAGGAGGTCAGTGTGTAGG -3'
5. 5'- GTGTGTTGAGCTTGGAGTTCGGG -3'
These sequences target different exons within human EGFR. Note that your selection from these sequences should be driven by desired outcomes. Some may need considerations such as possible pre-existing single nucleotide polymorphisms, the uniqueness of the sequence for the genome, and the efficiency of the particular sgRNA.
Please also remember that it will be essential to validate the on-target and off-target effects of your selected sgRNA in your specific experimental context. Additional steps may include sequencing and functional assays to confirm successful knockout.
There are online tools available such as Benchling, CHOPCHOP, and CRISPOR, which can assist you in designing sgRNA sequences for your specific needs.

\textbf{Note:}
The sequences provided by GPT4 did not align to the sequence of human EGFR sequences.

\subsubsection*{Example 5:}

\textbf{Prompt:} Could you give me the sgRNA sequence for CRISPR activation of human TGFBR1

\textbf{GPT4:}
To use CRISPRa (CRISPR activation) for gene activation, a series of sgRNAs that target various regions near the transcription start site of the gene of interest are typically used. Below is a set of TGFBR1 targeting sgRNA sequences found in the literature. These sgRNA sequences target the promoter of TGFBR1:

1. 5'- GAGTTAGGTTTCCCGTTGCG -3'
2. 5'- CAGGGTGGTGAGGCCATCTA -3'
3. 5'- GCCATCTAGTTCTCTCCGGA -3'
4. 5'- CAGACGTTGGAGGAGTTAGG -3'
5. 5'- TGACCACGGTCGCGCTCTCC -3'

In designing and choosing the most appropriate sgRNA, the use of software and online prediction tools, such as Benchling, Optimized CRISPR Design, and CHOPCHOP, can be highly beneficial.
Importantly, after performing CRISPRa, it's crucial to verify upregulation of TGFBR1 at both RNA (qRT-PCR) and protein (Western blot or ELISA) levels, and also monitor any change in cell phenotype. Off-target effects should also be evaluated. As with any experimental procedure, several control experiments, including a non-targeting sgRNA control, should be included.

\textbf{Note:}
The sequences provided by GPT4 did not align to the sequence of human TGFBR1 sequences.

\subsubsection*{Example 6:}

\textbf{Prompt:} Could you give me sequences of PCR primers to verify knockout of human EGFR using this guide ATACACCGTGCCGAACGCAC ?

\textbf{GPT4:}
To design PCR primers for verifying a knockout, you typically need to know the genomic context around your cut site which is located in the EGFR gene. 
The given gRNA sequence (ATACACCGTGCCGAACGCAC) targets human EGFR gene at the site 55,244,314 - 55,244,333 on chromosome 7 according to the Genome Reference Consortium Human Build 38 patch release 13 (GRCh38.p13). 
Based on this position, I suggest to use following primer sequences:
Forward primer: CAGGAAGGAAGGAGGAGGAAG
Reverse primer: GCAGCACGAGCTTAGTTAGG
These primers are designed for approximately 200 bp amplicon size and expected to perform reasonably well under standard PCR conditions. For best results, adjust conditions based on your specific PCR reagents.
The resulting PCR products should then be analyzed using methods such as Sanger sequencing or Next generation sequencing (NGS) to verify mutations indicative of knockout. However, please note that any online provided sequences should be further verified. If possible, consult with a geneticist/biologist/professional tool before utilization.

\textbf{Note:}
The primer sequences provided by GPT4 did not align to the sequence of human TGFBR1 sequences.

\subsection{Examples of incomplete information:}

\subsubsection*{Example 1:}

\textbf{Prompt:} How should I knock out human EGFR in my A549 cell line?

\textbf{GPT4:}
Here are the general steps to knockout the human EGFR gene in A549 cells using the CRISPR-Cas9 system:

1. gRNA Design: First, you have to design guide RNAs (gRNAs) that target the EGFR gene. There are many online tools available to assist you in designing gRNAs with high specificity and efficiency, like the CRISPOR and the Benchling tools.
2. Cloning: Clone the gRNAs into a plasmid that also expresses the Cas9 nuclease. This plasmid can either express the gRNA and Cas9 from separate promoters or express a single transcript that includes both the gRNA and Cas9 sequences.
2. Transfection: Deliver the plasmid containing the gRNA and Cas9 into the A549 cells. This can be achieved using a lipofection reagent such as Lipofectamine 3000, following the manufacturer's instructions.
3. Verification: After allowing time for the CRISPR-Cas9 system to induce double-strand breaks and errors in the DNA repair, which will result in insertions or deletions (indels) disrupting the EGFR gene, you have to verify the knockout. Genomic DNA PCR with subsequent Sanger sequencing or a mismatch detection assay like T7 endonuclease I or Surveyor nuclease can be used to confirm indels at the target site.
4. Functional Assays: This step involves verification of the knockout at the protein level. Western blotting or immunostaining against EGFR can be used to confirm reduced or absent protein level. Additionally, you may wish to determine the effects of the knockout on cell function, which might include effects on EGFR downstream signaling and cellular processes like proliferation and migration.
5. Validation: Validate the results using multiple clones and repeat the experiments to ensure the observed results are consistent and reproducible.
Note: It is very important to always include appropriate controls, such as untransfected cells or cells transfected with a non-targeting gRNA, in your experiments. Remember also to comply with your institutional or governmental regulations for working with CRISPR-Cas9 technology.

\textbf{Note:}
The answer is incomplete in multiple ways. 1. It does not tell the users to first select the CRISPR system. 2. It does not provide users with necessary sgRNA sequences. 3. It does not provide the protocol for the users to do the cloning and transfection. In addition, it provides repetitive information on verification but not the details.

\subsubsection*{Example 2:}
\textbf{Prompt:} How should I transfect the CRISPR knockout system into 293FT cells?

\textbf{GPT4: }
For transfection of 293FT cells with CRISPR knockout systems, lipofection is often used. Below are general steps:
1. Cell Culture: 293FT cells should be cultured under appropriate conditions (usually DMEM supplemented with 10
2. Plasmid DNA Preparation: Prepare your CRISPR plasmids that include Cas9 and your designed sgRNA. Make sure the plasmid DNA is pure and of high quality.
3. Transfection Using Lipofectamine: For 293FT cells, Lipofectamine 2000 or Lipofectamine 3000 is commonly used. Prepare the transfection complex by diluting the Lipofectamine reagent and plasmid DNA in Opti-MEM and combining them. Incubate the mixture for 5-20 minutes at room temperature to allow complex formation.
4. Transfection: Remove the growth medium from your cells, replace with fresh medium, and add the Lipofectamine-DNA complex to the cells. 
5. Post-transfection Culture: Incubate the cells at 37$^{\circ}$C in a CO2 incubator. Typically, gene editing is allowed to proceed for 48-72 hours before analysis.
6. Verification: After 48-72 hours, assess the transfection efficiency and gene editing effect. 
This is a general protocol and it's important that you optimize the conditions (like the ratio of Lipofectamine to DNA, incubation time, etc.) based on the specific requirements of your experiment. And always remember to include appropriate controls in your experiment to validate your results. 
Note: Follow all relevant institutional guidelines for genetic manipulation and cell culture to ensure biohazard safety.

\textbf{Note:}
The answer is incomplete as it did not provide the quantity of cells/plasmids/Lipofectamine reagents which are necessary to perform the experiment.

\subsubsection{Example 3:}

\textbf{Prompt:} How should I evaluate the CRISPRa activation efficiency of human EGFR in 293FT cells?

\textbf{GPT4: }
To evaluate the activation efficiency of CRISPRa (CRISPR activation) of the human EGFR gene in 293FT cells, you can use several methods:
1. Quantitative RT-PCR (qRT-PCR): This technology can be used to quantify the mRNA level of EGFR. If the CRISPRa system is working efficiently, you should observe an increase in EGFR mRNA levels compared to your control group.
2. Western Blot: You can also confirm the protein expression level of EGFR using Western blot after the gene has been activated. An efficient CRISPRa should result in an increase in EGFR protein levels.
3. Flow Cytometry or Immunofluorescence: These methods can be used to quantify and visualize the level of EGFR protein expression at single-cell resolution. 
4. Cellular Assays: Depending on the exact function of EGFR in your cells, you could use a variety of cellular assays (proliferation assays, wound healing assays, etc.) to observe the functional outcomes of the EGFR gene activation.
5. Next-Generation Sequencing: RNA-seq can be used not only to confirm the overexpression of EGFR but also to explore the downstream effects of its activation on a transcriptome-wide scale.
6. Off-target Analysis: Whole-genome sequencing or methods like Digenome-seq, GUIDE-seq, or CIRCLE-seq can be used to evaluate potential off-target gene activation by your CRISPRa system.
It's essential to include appropriate controls and perform experiments in replicates to ensure reliable results. Carry out normalization properly when quantifying the gene expression. Finally, it's recommended to confirm the activation efficiency with at least two independent methods such as qRT-PCR and Western blot.

\textbf{Note:}
The answer does not provide detailed information on how to perform the experiment using each of the method mentioned.

\section{Prompt Formats}
\label{sec:appendix:B}
\subsection{The prompt format of the task decomposition.}

We use the following prompt format for task decomposition for the LLM planner in the automation mode. The LLM planner interprets the user’s request and decomposes it into a list of tasks. The LLM planner is prompted to respect the task dependencies stated in the Task Description Table.

\begin{shaded}
\begin{lstlisting}
Please act as an expert in CRISPR technology. Given the user input, think step by step and generate a list of tasks for execution. First refer to the task description table below, and try to figure out if the user needs to directly jump into a task, or the user needs to complete several tasks. Make sure to respect the task depedencies and include all dependent tasks in the list.

Please format your response and make sure it is parsable by JSON.

## Task Description Table

{Task Description Table}

## Demonstrations:
If the user only needs to design guideRNA for knockout, then return ['knockout.StateStep1', 'knockout.StateStep3']. Reason: this directly matches knockout.StateStep3. But it needs to complete knockout.StateStep1 first, so both 'knockout.StateStep1' and 'knockout.StateStep3' are returned.

User Input:

"{user_message}"

Response format:

{{
"Thoughts": "<thoughts>",
"Tasks": ["<task1>", "<task2>"]  ## a list of task names 
}}
\end{lstlisting}
\end{shaded}

The task description table contains all the implemented tasks and their dependencies; see Table 1 for details.

\begin{shaded}
\begin{lstlisting}
For knockout

task name: task descriptions: dependency
knockout.StateStep1: Cas System selection for knockout : none
knockout.StateStep2: Delivery approach selection for knockout : none 
knockout.StateStep3: guideRNA design for knockout : needs to complete knockout.StateStep1 first
knockout.StateStep4: Experimental Protocol Selection for knockout : needs to complete knockout.StateStep2 first  
knockout.StateStep4_5_1_Sanger: Primer Design for knockout, Mutation sequencing by Sanger : none 
knockout.StateStep4_5_1_NGS: Primer Design for knockout, Mutation sequencing by next-generation sequencing (NGS) : none 

For base editing

task name: task descriptions: dependency
base_editing.StateStep1: Base Editor System selection for base editing : none
base_editing.StateStep2: guideRNA design for base editing : needs to complete base_editing.StateStep1 first
base_editing.StateStep3: Delivery approach selection for base editing : none 
base_editing.StateStep4: Experimental Protocol Selection for base editing : needs to complete base_editing.StateStep3 first 
base_editing.StateStep4_5_1_Sanger: Primer Design for base editing, Mutation sequencing by Sanger : none 
base_editing.StateStep4_5_1_NGS: Primer Design for base editing, Mutation sequencing by next-generation sequencing (NGS) : none 

For prime editing

task name: task descriptions: dependency
prime_editing.StateStep1: Prime Editing System selection for prime editing : none
prime_editing.StateStep2: Delivery approach selection for prime editing : none 
prime_editing.StateStep3: pegRNA design for prime editing : needs to complete prime_editing.StateStep1 first
prime_editing.StateStep4: Experimental Protocol Selection for prime editing : needs to complete prime_editing.StateStep2 first 
prime_editing.StateStep4_5_1_Sanger: Primer Design for prime editing, Mutation sequencing by Sanger : none 
prime_editing.StateStep4_5_1_NGS: Primer Design for prime editing, Mutation sequencing by next-generation sequencing (NGS) : none 

For CRISPRa/CRISPRi

task name: task descriptions: dependency
act_rep.StateStep1: Activation or repression system selection for CRISPRa/CRISPRi : none
act_rep.StateStep2: Delivery approach selection for CRISPRa/CRISPRi : none 
act_rep.StateStep3: guideRNA design for CRISPRa/CRISPRi : needs to complete act_rep.StateStep1 first
act_rep.StateStep4: Experimental Protocol Selection for CRISPRa/CRISPRi : needs to complete act_rep.StateStep2 first 
act_rep.StateStep4_5_1: Primer Design for CRISPRa/CRISPRi, qPCR : none 

For Off-Target Prediction

task name: task descriptions: dependency
off_target.StateStep1: Off-target search/predictiono using CRISPRitz: none
\end{lstlisting}
\end{shaded}

\subsection{The format of the prompt for LLM-agent.}

We synthesize the relevant information into \verb|{system_message}|, including the instruction of the current state, the interaction history between the agent and the system, and potentially the results from external tools and libraries. Next, we supply the meta request of the user in \verb|{meta_prompt}|. Then we prompt the LLM-agent to understand the current state and make decisions on behalf of the user. 
 
\begin{shaded}
\begin{lstlisting}
Please act as you are using the CRISPR design tool. Given the user meta request, the current inquiry provided by the tool, think step by step and generate an answer to the questions. Please format your response and make sure it is parsable by JSON.

Rules:

1. Answer the inquiry directly on behalf of the user. Don't raise any additional question to the user.
2. If the inquiry is a multiple-choice question, then directly output one choice. 
3. If the inquiry asks you to supply any gene sequence, then answer the question with "I don't know" and let the user take manual control.

User Meta Request:

"{meta_prompt}"

Current Inquiry:

"{system_message}"

Response format:

{{
"Thoughts": "<thoughts>",
"Answer": "<response string>"
}}


\end{lstlisting}
\end{shaded}

\section{The Rubrics for Human Evaluations}
\label{sec:appendix:C}
We evaluate the performance of CRISPR-GPT agent against ChatGPT v3.5 and ChatGPT v4 in four different aspects: Accuracy, Reasoning, Completeness and Conciseness. The detailed rubrics for expert evaluation are listed below.

\textbf{Accuracy}
\begin{itemize}[itemsep=1pt, parsep=1pt, topsep=2pt]
    \item 1 (Poor): The answer contains multiple factual errors or shows a misunderstanding of CRISPR technology.
    \item 2 (Fair): The answer has some correct elements but also includes significant inaccuracies that could lead to flawed experimental design if followed.
    \item 3 (Average): The answer is mostly accurate but may contain minor errors or oversights.
    \item 4 (Good): The answer is accurate, with only negligible errors that do not impact the overall validity of the information provided.
    \item 5 (Excellent): The answer is completely accurate, reflecting the current state of CRISPR research and methodologies.
\end{itemize}

\textbf{Reasoning}
\begin{itemize}[itemsep=1pt, parsep=1pt, topsep=2pt]
    \item 1 (Poor): The reasoning behind the answer is flawed or nonexistent; the logic is unclear or incorrect.
    \item 2 (Fair): The answer provides a rationale, but it is weak and may not support the conclusion or design effectively.
    \item 3 (Average): The answer's reasoning is solid for the most part, with some areas that could be better supported or explained.
    \item 4 (Good): The answer provides strong reasoning with clear and logical support for all claims and suggestions made.
    \item 5 (Excellent): The answer's reasoning is exceptional, providing insightful, well-supported explanations that enhance understanding of CRISPR designs.
\end{itemize}

\textbf{Completeness}
\begin{itemize}[itemsep=1pt, parsep=1pt, topsep=2pt]
    \item 1 (Poor): The answer is incomplete and lacks critical information required to form a complete understanding or action plan.
    \item 2 (Fair): The answer covers some necessary points but omits several important aspects that would be needed for a thorough CRISPR design.
    \item 3 (Average): The answer is fairly comprehensive but could be improved with additional details or coverage of more nuanced aspects of the design.
    \item 4 (Good): The answer is thorough, covering nearly all aspects required for a complete understanding and successful experimental setup.
    \item 5 (Excellent): The answer is entirely comprehensive, leaving no question unanswered and providing a full suite of information needed for CRISPR experimental design.
\end{itemize}

\textbf{Conciseness}
\begin{itemize}[itemsep=1pt, parsep=1pt, topsep=2pt]
    \item 1 (Poor): The answer is overly verbose and contains much irrelevant information, making it difficult to extract useful insights.
    \item 2 (Fair): The answer is longer than necessary with some extraneous content but still delivers a fair amount of relevant information.
    \item 3 (Average): The answer conveys the necessary information with some unnecessary detail but remains clear and understandable.
    \item 4 (Good): The answer is concise, with well-organized content that is directly relevant to the question asked, without any unnecessary information.
    \item 5 (Excellent): The answer is exceptionally concise, communicating the required information efficiently and effectively without any superfluity.
\end{itemize}

\begin{figure}[h]
    \includegraphics[width=0.01\textwidth]{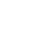} 
\end{figure}

\end{document}